\documentclass[11pt]{article}

\usepackage[final]{acl}

\usepackage{times}
\usepackage{latexsym}

\usepackage[T1]{fontenc}
\usepackage[utf8]{inputenc}

\usepackage{microtype}

\usepackage{inconsolata}

\usepackage{graphicx}
\graphicspath{ {./images/} {./latex/images/} }
\usepackage{amsmath}
\usepackage{amssymb}
\usepackage{todonotes}
\usepackage{booktabs}
\usepackage{multirow}

\title{APEX-MEM: Agentic Semi-Structured Memory with Temporal Reasoning for Long-Term Conversational AI}

\author{
  \textbf{Pratyay Banerjee},
  \textbf{Masud Moshtaghi},
  \textbf{Shivashankar Subramanian},
  \textbf{Amita Misra},
  \textbf{Ankit Chadha}
\\
\\
  Amazon, AGI, Sunnyvale, USA
\\
  \small{
    \texttt{\{pratyay, mmasud, ssangu, misrami, ankitrc\}@amazon.com}
  }
}

\begin{document}
\maketitle
\begin{abstract}
Large language models still struggle with reliable long-term conversational memory: simply enlarging context windows or applying naïve retrieval often introduces noise and destabilizes responses. We present APEX-MEM, a conversational memory system that combines three key innovations: (1) a property graph which uses domain-agnostic ontology to structure conversations as temporally grounded events in an entity-centric framework, (2) append-only storage that preserves the full temporal evolution of information, and (3) a multi-tool retrieval agent that understands and resolves conflicting or evolving information at query time, producing a compact and contextually relevant memory summary. This retrieval-time resolution preserves the full interaction history while suppressing irrelevant details. APEX-MEM achieves 88.88\% accuracy on LOCOMO's Question Answering task and 86.2\% on LongMemEval, outperforming state-of-the-art session-aware approaches and demonstrating that structured property graphs enable more temporally coherent long-term conversational reasoning.

\end{abstract}

\section{Introduction}

Large language models (LLMs) have demonstrated remarkable capabilities in understanding and generating human-like text, yet they fundamentally struggle with maintaining coherent memory across extended conversations. In applications such as personal assistants and complex goal-oriented tasks, users expect systems to remember prior discussions, accumulate knowledge across sessions, and adapt responses based on evolving context. This expectation is magnified in open-domain conversational agents, which must sustain far more  turns across diverse topics \cite{kim-etal-2023-aligning}. 

A naïve solution is to leverage models with larger context windows to retain longer conversational histories. However, longer contexts trade memory for noise, severely increasing the risk of irrelevant facts or hallucinations in the output of LLMs \cite{maharana2024locomo, zhang2024chain, du2025context}. To reduce noise, researchers started leveraging retrieval augmented generation (RAG) techniques to identify most relevant parts of conversation~\cite{packer2023memgpt, lee2024readagent, alonso2024}. In this approach, textual segments or summaries are stored and retrieved during conversation. Yet these methods also suffer from fundamental limitations: retrieval accuracy does not reliably translate to answer accuracy when summaries lose critical details, and increasing the number of retrieved segments frequently reintroduces noise \cite{ICLR2025_e56f394b, zhang2024chain}. More broadly, RAG treats memory as unstructured text and offers no mechanism to maintain canonical entities, track evolving facts, or distinguish persistent information from ephemeral conversational content. As conversational histories grow, LLMs struggle to maintain factual consistency, entity continuity, and temporal coherence~\cite{byerly2024self}.

These limitations motivate the shift toward structured memory representations, which explicitly organize conversational knowledge to reduce noise, improve retrieval precision, preserve coherence across long-term interactions, and enable relational and temporal reasoning that unstructured text memory cannot support. Recent research has explored this direction through various architectures: Mem0$^{g}$ represents memory as entity-centric graphs to capture relational structure~\cite{chhikara2025mem0}, A-MEM constructs dynamic knowledge networks inspired by strategic note taking technique (Zettelkasten) to maintain long-term memory consistency~\cite{xu2025mem}, Zep builds temporally-aware knowledge graphs to track evolving facts across sessions~\cite{rasmussen2025zep}, and Semantic Anchoring leverages linguistic structure and entity-aware memory to enhance persistence and coherence in conversational reasoning~\cite{chatterjee2025semantic}. These approaches demonstrate that structured memory provides a principled foundation for improving long-term conversational performance beyond what search-based unstructured text memory can achieve. Despite these advances, current memory systems face two critical limitations. First, entity-centric graph approaches like Mem0, have limited entity classes and store information primarily as relationships between entities, limiting their ability to capture nuanced attributes and temporal evolution of facts. Second, existing systems consolidate (or overwrite previous) information which risks losing important contextual details needed for temporal reasoning and conflict resolution. 

To overcome the limitations of existing structured memory systems, we present APEX-MEM, a conversational memory framework that combines property graphs supported by a domain-agnostic ontology, append-only event storage, and retrieval-time temporal resolution. Our contributions are threefold: 1) We introduce a hybrid entity-event ontology for conversational memory that combines entity-centric and event-centric temporal modeling. Unlike purely entity-focused approaches that struggle with temporal evolution, our ontology represents conversational events as first-class citizens enabling fine-grained temporal reasoning while maintaining entity coherence. 2) To avoid the risk of losing information, we use append-only event storage where facts are anchored to temporally grounded events rather than directly to entities. This preserves the full evolution of information including contradictions and revisions enabling retrieval-time resolution based on temporal validity rather than premature commitment to a single current state. 3) We develop a complementary multi-tool retrieval framework that combines entity linking (\textsc{EntityLookup}), structured graph traversal (\textsc{GraphSql}), and hybrid (semantic+lexical) search (\textsc{Search}), all supported by meta-level planning guidance tool (\textsc{SchemaViewer}). \textsc{EntityLookup} provides entity-centered access and canonicalization, enabling resolution of surface mentions to canonical graph identifiers. \textsc{GraphSql} enables structured temporal reasoning through SQL-based graph traversal, supporting fact evolution tracking, temporal ordering via validity intervals, duration calculations, and multi-hop relationship traversal across events. $\textsc{Search}$ enables flexible hybrid retrieval across the entire graph memory to retrieve high-relevance subgraphs especially for open-domain questions.

\begin{figure*}[t]
  \centering
 \includegraphics[width=\textwidth,height=0.38\textheight,keepaspectratio]{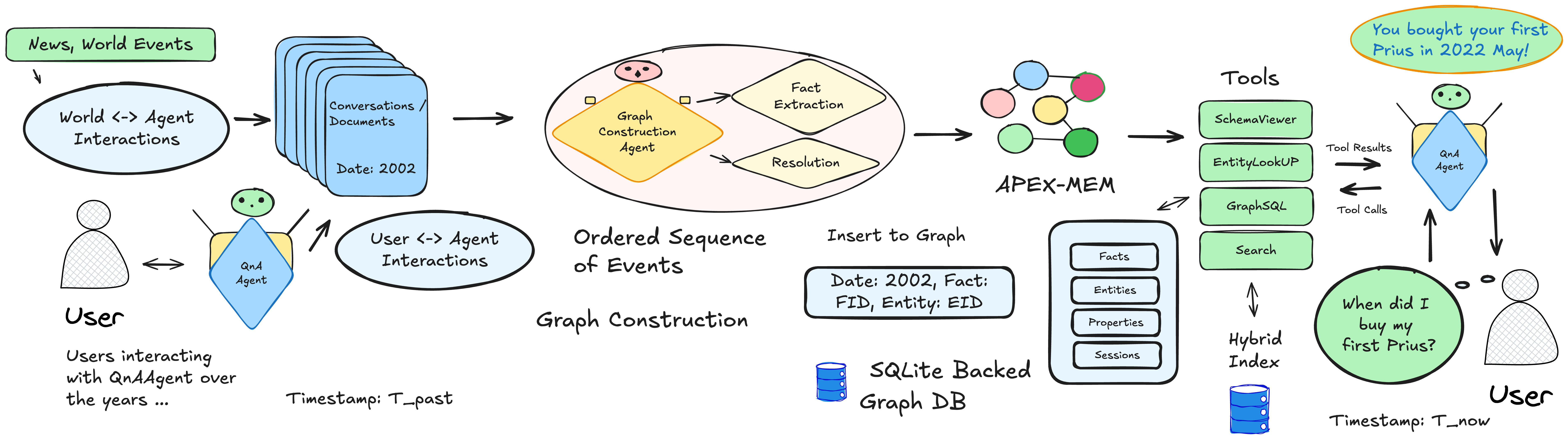}
\caption{End-to-end pipeline for constructing and querying APEX-MEM Graph, showing data flow from unstructured conversation to GraphSQL QA Agent.}
\end{figure*}

%In this paper, we introduce a novel approach to conversational memory through ontology-supported property graphs with append-only semantics and retrieval-time resolution. Unlike entity-centric relational graphs in Mem0, our system constructs rich property graphs where information is stored as detailed attributes on nodes and edges, supported by a domain-agnostic ontology that provides structural consistency without limiting expressiveness. We adopt an append-only architecture at graph construction time—when memory updates occur, we preserve both the original and new information as separate events in the graph, deferring resolution to retrieval time. This design enables the generation agent to access the full temporal context of evolving information and perform intelligent resolution based on query-specific requirements, rather than making premature decisions about which information to retain during ingestion. The ontology provides semantic grounding for entity linking and enables sophisticated graph traversal through Graph SQL access patterns, allowing a graph agent to efficiently retrieve relevant context while maintaining the full richness of conversational history.

\section{Related Work}
\textbf{Early Approaches to Conversational Memory.} Initial efforts to extend conversational reasoning in LLMs relied on larger context windows or retrieval-augmented generation (RAG). The LOCOMO~\cite{maharana2024locomo} showed inherent limitations in both approaches due to added noise in the context window: while GPT-4-Turbo achieved 51.6\% F1 with 128K context vs 35.9\% for GPT-3.5-Turbo at 16K, adversarial performance dropped sharply to 15.7\% F1 as models attended to irrelevant details.  Hence RAG methods only offered modest gains (GPT-3.5 + RAG reaching 43.3\% F1 vs 22.4\% no retrieval)~\cite{zhang2024chain}.

\textbf{First-Generation Memory Systems.} Researchers introduced explicit memory management architectures to address these weaknesses. MemGPT~\cite{packer2023memgpt} pioneered an OS-inspired memory hierarchy (26.65\% F1 on LOCOMO). ReadAgent~\cite{lee2024readagent} used paginated "gist memory" effective for narrative QA but achieved only 9.15\% F1 on conversations. MemoryBank~\cite{zhong2023memorybank} adopted psychologically motivated updates but struggled with factual retention (5.0\% F1). OpenAI's Memory feature performed comparably to GPT-4-Turbo (52.9\%), indicating general-purpose memory layers remained insufficient. These systems demonstrated that architectural complexity alone was inadequate without mechanisms for preserving information fidelity and mitigating noise accumulation.

\textbf{Advanced Memory Architectures.} Recent work introduced sophisticated architectures for improved retrieval and reasoning. A-MEM~\cite{xu2025amem} proposed agentic, Zettelkasten-inspired memory with autonomous linking (27.0\% F1 single-hop). H-MEM~\cite{sun2025hmem} adopted hierarchical retrieval, reporting +21-point gains on multi-hop tasks. Mem0~\cite{chhikara2025mem0} implemented entity-centric relational graphs achieving 67.1\% single-hop and 51.2\% multi-hop accuracy with 91\% latency reduction; Mem0$^{g}$ improved temporal reasoning (58.1\%) but lacked rich property attributes. Semantic Anchoring~\cite{chatterjee2025semantic} demonstrated benefits of linguistic structure for factual stability. Zep builds temporally-aware knowledge graphs to track evolving facts across sessions \cite{rasmussen2025zep}. However, it relies heavily on text retrieval technique to search context. MIRIX~\cite{wang2025mirix} achieved state-of-the-art 85.4\% accuracy using six specialized memory stores with multi-agent routing, though its complexity and eager-update strategy risked losing nuanced temporal information. These systems show benefits of structured memory and bring up trade-offs in complexity, expressiveness, and temporal coherence preservation.

%Recent work explored more sophisticated memory organizations. A-MEM~\cite{xu2025amem} introduced agentic memory with autonomous linking, achieving strong multi-hop reasoning but modest absolute scores (27.0\% F1 single-hop). H-MEM~\cite{sun2025hmem} proposed hierarchical memory with layer-by-layer retrieval, reportedly improving multi-hop reasoning by +21 F1 points while reducing search costs. However, these systems introduced significant complexity without fully bridging the gap to human-level performance (87.9\% F1).

%\textbf{State-of-the-Art Memory Systems} Two recent systems have pushed performance to new levels. Mem0~\cite{chhikara2025mem0} introduced a two-phase pipeline with entity-centric relational graphs, achieving 67.1\% on single-hop and 51.2\% on multi-hop queries with 91\% latency reduction. Its graph variant Mem0$^g$ further improved temporal reasoning (58.1\%) but entities with typed relationships rather than rich property attributes. MIRIX~\cite{wang2025mirix} employed six specialized memory stores with multi-agent routing, achieving 85.4\% accuracy—the previous state-of-the-art. However, MIRIX's complexity raises practical deployment concerns. Both systems use eager update strategies that potentially lose contextual details needed for temporal reasoning. Our work addresses these limitations through ontology-supported property graphs with append-only architecture and retrieval-time resolution.

\section{APEX-MEM Graph Construction}

APEX-MEM is a directed property graph $G = (V,E,\Pi,\Lambda)$ where: $V$  is the set of nodes (entities, events, facts, etc.), $E \subseteq V\times V$ is the set of labeled, typed edges, $\Pi : V \cup E \to \mathcal{P}(K\times S)$ maps each element to a set of key–value property pairs, $ \Lambda : V \cup E \to T$ assigns a type (from the ontology) to every node and edge.

For every source document ($d_i \in D$) we induce a sub‑graph ($g_i$).  Merging is performed incrementally using a \textit{soft‑canonicalization} function:
\begin{equation}
\begin{split}
G^{(t+1)} ;\leftarrow; \mathrm{Merge}\bigl(G^{(t)}, g_{t}\bigr) \\ 
    \mathrm{Merge}(G,g) := (V\cup V_g,; E\cup E_g,; \ldots)
\end{split}
\end{equation}

\noindent where candidate entity nodes and properties are fused when corresponding criteria for Entity and Property Resolution is met. Property graphs allow rich key-value representation of structured information about the entity and how it is referred in the conversation, in contrast to Mem0 where conversation is not ontologized. 

\subsection{Ontology}
We propose a temporal event ontology for conversational memory systems grounded in world knowledge representation. The ontology defines a comprehensive entity type hierarchy $\tau \in \mathcal{T}$ comprising 35 classes analogous to YAGO taxonomies~\cite{yago2024}, spanning agents (Person, Organization, Corporation), living organisms (Animal, Plant, Taxonomy), spatiotemporal constructs (Place, Event, Time), physical artifacts (Product, Device, Vehicle), digital objects (Software, Dataset, Service), information resources (CreativeWork, Document, Message), financial instruments (Stock, Contract), health concepts (Food, Medication, Disease), and abstract entities (Topic, Metric, Task). Each entity $e = (n, \tau, \rho, \text{id})$ is characterized by name $n$, type $\tau$, conversational role $\rho \in \mathcal{R}$ (Speaker, Listener, Agent, Mentioned), and optional external identifier. Facts are represented as temporally-grounded subject-property-value assertions $f = (s, p, v, \delta, [t_{\text{from}}, t_{\text{to}}], c, \mathcal{E})$ where $s$ denotes the subject entity, $p$ the property, $v$ the value with data type $\delta$, $[t_{\text{from}}, t_{\text{to}}]$ the temporal validity interval, $c \in [0,1]$ the confidence score, and $\mathcal{E}$ the evidence set. All facts are anchored to conversational events $\varepsilon = (\text{type}, T, L, P, F, \mathcal{E}_\varepsilon)$ where $T$ represents the event timestamp, $L$ the location, $P \subseteq \{e_i\}$ the participant set, $F \subseteq \{f_j\}$ the associated facts, and $\mathcal{E}_\varepsilon$ the supporting textual evidence, enabling temporal reasoning over evolving world knowledge extracted from dialogue. This ontology allows flexible property attachment beyond strict schema constraints, enabling domain adaptation while maintaining structural consistency.

\subsection{Entity and Property Resolution}
To ensure robust entity and property canonicalization across conversational turns we resolve entities and properties in the conversation text against existing set of entities. This maintains provenance through confidence scoring and explicit rationale generation.
We use a retrieval-augmented generation approach combining dense semantic search over known entities with structured LLM reasoning. Given a mention $m$ in conversational text, the entity resolver retrieves top-$k$ candidates $C = \{c_1, \ldots, c_k\}$ from a dense vector index using cosine similarity, where each candidate $c_i = (\text{id}_i, \text{text}_i, s_i)$ includes an identifier, textual representation, and similarity score $s_i \in [0,1]$. A structured LLM then evaluates candidates against the mention and contextual information to produce a resolution decision $d \in \{\text{choose\_existing}, \text{propose\_new}, \text{none}\}$ with output $o = (d, \text{id}, n_{\text{norm}}, \tau, A, c, r)$ where $\text{id}$ is the resolved or newly generated entity identifier, $n_{\text{norm}}$ the normalized name, $\tau$ the entity type, $A$ a set of aliases, $c \in [0,1]$ the confidence score, and $r$ the rationale. Property resolution follows an analogous pipeline, additionally normalizing property names to snake\_case and inferring data types $\delta \in \{\text{str}, \text{int}, \text{float}, \text{bool}, \text{date}, \text{datetime}, \text{enum}, \text{url}, \text{list}\}$.

\subsection{Fact Extraction}
Structured fact extraction is performed through few-shot prompted large language models with schema-constrained generation. Given a conversational turn $u = (s, l, \text{text}, t_{\text{anchor}}, \text{ctx})$ where $s$ denotes the speaker, $l$ the listener, $t_{\text{anchor}}$ the anchor timestamp (the conversation temporal timestamp), and $\text{ctx}$ the recent conversational context, the extractor generates a structured event representation $\varepsilon = (\text{type}, T, L, P, F, \mathcal{E})$ conforming to a predefined schema, defined in the Ontology. The extraction prompt comprises hand-crafted high-quality few-shot exemplars demonstrating comprehensive fact extraction across diverse conversational patterns, including factual assertions, numerical data, emotional states, environmental conditions, and personal attributes. The LLM output is validated, with strict type specifications: participants must conform to $\tau \in \mathcal{T}$ and $\rho \in \mathcal{R}$, facts must specify data types $\delta \in \Delta$, and temporal expressions are normalized to ISO 8601 format relative to $t_{\text{anchor}}$. All facts include confidence scores $c \in [0,1]$ and supporting evidence $\mathcal{E} = \{(e_{\text{text}}, e_{\text{turn}}, e_{\text{span}})\}$ linking assertions to source utterances. This structured approach ensures consistent, machine-readable knowledge extraction while preserving provenance and enabling downstream temporal reasoning over evolving conversational state.

\section{Graph Agents}
%We use ReAct (Reason + Act) framework~\cite{yao2022react} to build our graph agents for accessing the relevant information in APEX-MEM graph. Given an input question $x$ and interaction history $h_t$, a ReAct agent implements a policy $\pi_\theta$ parameterized by an LLM that at each step $t$ samples a thought $r_t \in \mathcal{R}$ and an action $a_t \in \mathcal{A}$, i.e.\ $(r_t,a_t) \sim \pi_\theta(\cdot \mid x,h_t)$, where actions correspond to calls to external knowledge sources (e.g., $\textsc{Search}(q')$, $\textsc{EntityLookup}(p)$). The environment executes $a_t$ and returns an observation $o_t \in \mathcal{O}$, and the history is updated as $h_{t+1} = h_t \cup \{(r_t,a_t,o_t)\}$; after $T$ steps the model emits a final answer $y$ conditioned on $(x,h_T)$.

%\subsection{Graph Retrieval Tools}
APEX-MEM Graph QnA agent is a ReAct-style agent~\cite{yao2022react} operating over a temporal property-graph database $\mathcal{G}$ instantiated by the SQLite schema in \texttt{DB\_SCHEMA} (tables \texttt{entities}, \texttt{properties}, \texttt{facts}, \texttt{events}, \texttt{evidence}, \texttt{turns} and their lexical search views). Given a natural-language question $x$ and interaction history $h_t$, the agent implements a policy $\pi_\theta$ that, at step $t$, generates a reasoning trace $r_t$ and selects an action $a_t \in \mathcal{A}$,
\[
(r_t,a_t) \sim \pi_\theta(\cdot \mid x, h_t).
\]
Actions are either (i) tool invocations $a_t = (T_t, z_t)$, where $T_t \in$ $\{\textsc{SchemaViewer},$ $\textsc{EntityLookUp},$ $\textsc{GraphSQL},$ $\textsc{Search}\}$ and $z_t$ are structured arguments conforming to the corresponding tool inputs, or (ii) a special \textsc{Answer} action that emits a final answer $y$. The agent resolves temporal references in each turn into dates and date-ranges which is used when a tool is invoked. Tool outputs $o_t$ are appended to the history, $h_{t+1} = h_t \cup \{(r_t, a_t, o_t)\}$, enabling multi-step reasoning that interleaves natural-language planning with structured access to entities, properties, events, evidence, and conversation turns stored in $\mathcal{G}$ via the tools exposed by \texttt{QnAAgent}.

\subsection{\textsc{SchemaViewer}}
The \textsc{SchemaViewer} tool is a schema- and strategy-inspection operator
\begin{equation}
    T_{\text{schema}} : \{0,1\}^2 \rightarrow \mathcal{S},
\end{equation}
which, given boolean flags $(b_{\text{ex}}, b_{\text{guide}})$ indicating whether to include examples and a usage guide, returns a structured schema view $s \in \mathcal{S}$. The agent uses \textsc{SchemaViewer} both to inspect the relational schema of the graph-backed SQLite database and to obtain query and tool-usage recommendations (e.g., when to call entity, event, instructions on how to do temporal reasoning using SQL), effectively acting as a meta-level planner aid for Graph QnA.

\subsection{\textsc{EntityLookup}}
The \textsc{EntityLookUp} tool is a entity-retrieval operator
\begin{equation}
    T_{\text{ent}} : \mathcal{Q} \times \mathbb{N} \rightarrow \mathcal{D}_{\text{ent}},
\end{equation}

mapping a free-text query $q \in \mathcal{Q}$ and a top-$k$ budget $K$ to a ranked list of entity documents $(d_1,\dots,d_K) \in \mathcal{D}_{\text{ent}}$. It first retrieves candidate entity ids from a hybrid index (combining dense and lexical search), then queries the underlying \texttt{GraphDB} over \texttt{entities}, \texttt{facts}, \texttt{properties}, \texttt{events}, and \texttt{evidence} to construct, for each entity id $e$, a document
$d = (\textit{id},\,$$ \textit{name},\, \textit{type},\, $$\textit{latest},\, \textit{anchors},\, $$\textit{last\_anchor},\, $$ \textit{facts}),$
where \textit{latest} and \textit{facts} are markdown tables summarizing recent property values and \textit{anchors}/\textit{last\_anchor} expose temporal context via \textit{events.anchor\_datetime}. The agent uses \textsc{EntityLookUp} to canonicalize surface forms to graph ids and to retrieve time-aware fact snapshots that ground downstream reasoning and SQL queries.

\subsection{\textsc{GraphSQL}}
The \textsc{GraphSQL} tool is a read-only SQL interface
\[
T_{\text{sql}} : \mathcal{S}_{\text{sql}} \times \mathcal{P}_{\text{sql}} \rightarrow \mathcal{R}_{\text{sql}},
\]
where $\mathcal{S}_{\text{sql}}$ is the set of safe SQLite \emph{SELECT} (or \emph{WITH\,$\dots$\,SELECT}) statements over the whitelisted tables: \texttt{events}, \texttt{facts}, \texttt{evidence}, \texttt{entities}, \texttt{event\_participants}, \texttt{properties}, \texttt{turns}. 
$\mathcal{P}_{\text{sql}}$ is the space of named-parameter maps, and $\mathcal{R}_{\text{sql}}$ is the space of result tables. The tool first validates the statement, enforcing a single read-only statement and forbidding Updates, or DDL then executes it against the \texttt{GraphDB} and returns sql outputs wrapped as markdown. \textsc{GraphSQL} is invoked when the agent needs precise graph reasoning (e.g., joining entities and events), aggregations, mathematical, or temporal computations based on \texttt{anchor\_datetime}, going beyond what pure retrieval can provide.

\subsection{\textsc{Search}}
The \textsc{Search} tool conceptually exposes a hybrid retrieval layer over the graph, its relational views, and semantic indices, mapping a query $q$ to a composite context
\[
T_{\text{search}} : \mathcal{Q} \rightarrow \mathcal{C}, \quad T_{\text{search}}(q) = (E_q, P_q, \mathcal{V}_q, \mathcal{T}_q),
\]
where $E_q$ are candidate entities, $P_q$ are candidate properties, $\mathcal{V}_q$ are candidate events/evidence, and $\mathcal{T}_q$ are relevant conversation turns. From the agent's perspective, \textsc{Search} provides a unified, hybrid graph--entity--property--SQL and semantic search capability: it retrieves a high-relevance subgraph around the question, which can then be further filtered or aggregated via \textsc{GraphSQL} before the ReAct policy emits the final answer.

\section{APEX-MEM Online Construction}
For very-long conversations, where the number of documents is a magnitude higher  $>10^3$, and its infeasible and unnecessary to construct a complete APEX-MEM Graph offline, as significant set of conversations or documents are irrelevant to expected user questions. For such cases, we construct APEX-MEM online, where, given a set of Documents ($D$) and input question ($Q$), we determine document relevance using semantic and lexical search, and limit Graph construction to those sub-set of temporally-ordered Documents $D_{rel}$, with $Relevance(d_i | Q) > \Theta_{rel}$.

\begin{table*}[t]
\centering
\scriptsize
\setlength{\tabcolsep}{1.5pt}
\begin{tabular}{@{}cccccccccc@{}}
\toprule
\multicolumn{1}{l}{\textbf{Method}}  & \multicolumn{1}{l}{}             & \multicolumn{1}{l}{} & \multicolumn{1}{l}{\textbf{Single-Hop}} & \multicolumn{1}{l}{\textbf{Multi-Hop}} & \multicolumn{1}{l}{\textbf{Temporal}} & \multicolumn{1}{l}{\textbf{Open-Domain}} & \multicolumn{1}{l}{\textbf{Adversarial}} & \multicolumn{1}{l}{\textbf{Overall}} & \multicolumn{1}{l}{\textbf{w/o Adv.}} \\ \midrule
\multirow{6}{*}{APEX-MEM}   & \textbf{Graph}                            & \textbf{QA Agent }            & \multicolumn{1}{r}{}           & \multicolumn{1}{r}{}          & \multicolumn{1}{r}{}         & \multicolumn{1}{r}{}            & \multicolumn{1}{r}{}            & \multicolumn{1}{r}{}        & \multicolumn{1}{r}{}        \\ \cmidrule(lr){2-3}
                            & APEX-MEM                         & Claude 4.5 Haiku     & 85.46\%                        & 84.74\%                       & 79.17\%                      & 89.18\%                         & \textbf{87.22}\%                         & 84.92\%                     & 84.25\%                     \\
                            & APEX-MEM                         & Claude 4.5 Sonnet    & \textbf{89.36}\%                        & \textbf{86.92}\%                       & \textbf{90.63}\%                      & 87.75\%                         & 86.10\%                         & 88.41\%                     & 89.08\%                     \\
                            & APEX-MEM                         & Claude 3.5 Sonnet    & 87.58\%                        & 84.74\%                       & 87.50\%                      & 88.94\%                         & 86.10\%                         & 86.90\%                     & 87.13\%                     \\
                            & APEX-MEM                         & GPT5                 & 89.88\%                        & 86.29\%                       & \textbf{90.63}\%                      & \textbf{91.68}\%                         & 86.77\%                         & \textbf{88.88}\%                     & \textbf{89.49}\%                     \\
                            & APEX-MEM                         & GPT4o                & 88.47\%                        & 85.46\%                       & 83.49\%                      & 86.46\%                         & 84.98\%                         & 86.35\%                     & 86.75\%                     \\ \midrule
\multirow{5}{*}{LOCOMO Baselines}  & \textbf{Retrieval}                        & \textbf{Agent}                &                                &                               &                              &                                 &                                 &                             &                             \\ \cmidrule(lr){2-3}
                            & SimpleSearch                     & GPT5                 & 76.60\%                        & 71.96\%                       & 72.92\%                      & 83.95\%                         & 72.65\%                         & 77.90\%                     &                             \\
                            & SimpleSearch                     & GPT4                 & 46.10\%                        & 51.40\%                       & 37.50\%                      & 73.60\%                         & 57.17\%                         & 60.67\%                     &                             \\
                            & SimpleSearch                     & Claude 4.5 Sonnet           & 71.99\%                        & 73.52\%                       & 73.96\%                      & 81.33\%                         & 74.22\%                         & 76.79\%                     &                             \\
                            & SimpleSearch                     & Claude 3.5 Sonnet          & 55.32\%                        & 61.99\%                       & 61.46\%                      & 79.55\%                         & 72.87\%                         & 70.90\%                     &                             \\ \midrule
\multirow{4}{*}{MemInsight} & MemInsight + Attribute Retrieval & Claude 3.5 Sonnet               & 74.82\%                        & 74.45\%                       & 72.92\%                      & 80.02\%                         & 78.92\%                         & 77.79\%                     &                             \\
                            & MemInsight + RAG Baseline        & Claude 3.5 Sonnet                & 79.43\%                        & 82.55\%                       & 77.08\%                      & 75.15\%                         & 79.15\%                         & 77.95\%                     &                             \\
                            & MemInsight                       & LLamaV3              & 76.95\%                        & 81.31\%                       & 80.21\%                      & 76.93\%                         & 78.92\%                         & 78.25\%                     &                             \\
                            & MemInsight                       & Mistral              & 74.47\%                        & 82.55\%                       & 81.25\%                      & 78.60\%                         & 81.17\%                         & 79.36\%                     &                             \\ \midrule
\multirow{8}{*}{Memory}     & \multicolumn{2}{c}{\textbf{Memory Agent}}                        & \multicolumn{1}{l}{}           & \multicolumn{1}{l}{}          & \multicolumn{1}{l}{}         & \multicolumn{1}{l}{}            & \multicolumn{1}{l}{}            & \multicolumn{1}{l}{}        & \multicolumn{1}{l}{}        \\ \cmidrule(lr){2-3}
                            & \multicolumn{2}{c}{OpenAI}                              & 63.79\%                        & 42.92\%                       & 62.29\%                      & 21.71\%                         & N/A                             & 52.90\%                     & 52.90\%                     \\
                            & \multicolumn{2}{c}{AMEM}                               & 39.79\%                        & 18.85\%                       & 54.05\%                      & 49.91\%                         & N/A                             & 48.38\%                     & 48.38\%                     \\
                            & \multicolumn{2}{c}{Zep}                                 & 61.70\%                        & 41.35\%                       & 76.60\%                      & 49.31\%                         & N/A                             & 75.14\%                     & 75.14\%                     \\
                            & \multicolumn{2}{c}{MemGPT/LangMem}                      & 62.23\%                        & 47.92\%                       & 71.12\%                      & 23.43\%                         & N/A                             & 58.10\%                     & 58.10\%                     \\
                            & \multicolumn{2}{c}{Memory-R1}                           & 59.83\%                        & 53.01\%                       & 68.78\%                      & 51.55\%                         & N/A                             & N/A                         & N/A                         \\
                            & \multicolumn{2}{c}{Mem0}                                & 65.71\%                        & 47.19\%                       & 75.71\%                      & 58.13\%                         & N/A                             & 68.44\%                     & 68.44\%                     \\
                            & \multicolumn{2}{c}{Mnemosyne}                           & 62.78\%                        & 49.53\%                       & 60.42\%                      & 53.03\%                         & N/A                             & N/A                         & N/A                         \\
                            & \multicolumn{2}{c}{Nemori}                           & 84.9\%                        & 75.1\%                       & 77.60\%                      & 51.0\%                         & N/A                             & 79.4\%                      & 79.4\%                      \\
                            & \multicolumn{2}{c}{MIRIX}                           & 85.11\%                        & 83.70\%                       & 65.62\%                      & 88.39\%                         & N/A                             & 85.38\%                     & 85.38\%                     \\ \midrule
\multirow{2}{*}{Full Context}     & \multicolumn{2}{c}{\textbf{QA Agent}}                        & \multicolumn{1}{l}{}           & \multicolumn{1}{l}{}          & \multicolumn{1}{l}{}         & \multicolumn{1}{l}{}            & \multicolumn{1}{l}{}            & \multicolumn{1}{l}{}        & \multicolumn{1}{l}{}        \\ \cmidrule(lr){2-3}
                                & \multicolumn{2}{c}{GPT4o}                           & 88.53\%                        & 77.70\%                       & 71.88\%                      & 92.70\%                         & N/A                             & 87.52\%                     & 87.52\%                     \\

\bottomrule
\end{tabular}
\caption{LOCOMO Category Type Evaluation Results}
\label{tab:locomo_evals}
\end{table*}

\section{Experiments and Analysis}
We evaluate APEX-MEM on the following different datasets that require diverse reasoning, signal-to-noise ratio, and are challenging to LLMs. We try to answer the following research questions:
RQ1. How does APEX-MEM compare to existing state-of-the-art for Memory-based Tasks?
RQ2. How important is each tool for APEX-MEM-based QnAAgent ?
RQ3. Does APEX-MEM improve QA Agent's ability to handle complex scenarios compared to other Deep research agents?
RQ4. How well does APEX-MEM generalize?

\begin{table}[ht]
\centering
\small
\setlength{\tabcolsep}{4pt}
\begin{tabular}{@{}lccc@{}}
\toprule
\textbf{Model} & \textbf{Fact} & \textbf{Schema} & \textbf{Entity/Prop.} \\
               & \textbf{Extraction} & \textbf{Coverage} & \textbf{Resolution} \\ \midrule
GPT4o             & 94.2\%                     & 75.7\%                      & 98.1\%                        \\
Claude Sonnet 4.5 & 97.3\%                   & 91.1\%                      & 98.2\%                        \\
Claude Haiku 4.5  & 95.8\%                      & 90.3\%                      & 95.4\%                        \\
Qwen3-14B         & 95.4\%                      & 88.9\%                      & 92.5\%                        \\ \bottomrule
\end{tabular}
\caption{APEX-MEM Construction metrics for fact extraction (precision of extracted facts), schema coverage (plausible properties covered), and entity resolution (detecting and linking to proper entity). We measure these metrics with 500 Random turns from LoCoMo and LongMemEval. GPT5 is used as the judge.}
\label{tab:apex-mem-const}
\end{table}

\subsection{Datasets}
\paragraph{LOCOMO} (Long-term Conversational Memory) is a benchmark for evaluating agent memory over extended multi-session dialogues spanning weeks-long interactions with evolving user preferences, personal facts, and temporal events~\cite{maharana2024locomo}. Its evaluation targets long-term recall, relevance discrimination, and cross-session consistency. Questions are categorized as single-hop, multi-hop, temporal, open-domain, and adversarial (unanswerable). Following \citet{chhikara2025mem0}, we use LLM-as-a-Judge to assess factual accuracy, relevance, completeness, and contextual appropriateness of generated answers against ground truth.

\paragraph{LongMemEval} examines LLM ability to process and reason over extremely long inputs including multi-document collections, extended narratives, and dense conversational histories~\cite{wu2024longmemeval}. The benchmark emphasizes context-length generalization, factual recall, cross-episode reasoning, and robustness under large-scale sequences, providing a standardized framework for evaluating long-context architectures. We measure performance using the recommended LLM-as-a-Judge for answer quality and factuality scores.

\paragraph{SealQA-Hard} challenges search-augmented LLMs on fact-seeking questions where web search yields conflicting or noisy results. We use SEAL-HARD to assess factual accuracy and reasoning on questions where chat models (e.g., GPT-4.1) achieve near-zero accuracy. SEALQA presents long-context, multi-document "needle-in-a-haystack" scenarios with 30 web-retrieved documents containing 1/2 gold documents at unknown positions. To align with memory benchmarks, we order documents by published time as observation time, processing each as an agent-world interaction. Each question-search pair constitutes a separate session. Performance is measured using LLM-as-a-Judge for answer quality and factuality.

\subsection{Experimental Setup}

For APEX-MEM construction task, refer to Table \ref{tab:apex-mem-const}, for how different LLMs perform at Fact Extraction and Entity Resolution. We construct APEX-MEM using Claude Sonnet 4.5 for Fact Extraction and Claude Haiku 4.5 for Entity and Property Resolution, to balance cost versus task performance.
We test different LLMs (Claude, GPT) for QnA Agents. 
All Tools are used with a max limit of 40 for ReACT tool invocations. We adopt APEX-MEM Online for LongMemEval and SealQA, extracting Entities and Facts from conversations which are marked relevant with $\Theta_{rel}> 0.2$. For LOCOMO, we construct a APEX-MEM for all input sessions in respective conversations. We re-implement LOCOMO baselines to measure the impact of new versions of Claude and GPTs. For LongMemEval we implement a stronger Search baseline of a Expanded Sessions that include top-5 relevant Sessions.
%Keep this required as part of approval
For all other benchmarks we used reported numbers from~\cite{salama2025,pham2025sealqa,wang2025mirix}. To make the results replicable, we set the temperature to 0 wherever applicable.  For each LLM-as-a-Judge, we report mean of 3 trials, with $<\pm1$ standard deviation.

\begin{table*}[t]
\resizebox{\linewidth}{!}{
\begin{tabular}{@{}cccccccc@{}}
\toprule
\multicolumn{2}{c}{\textbf{Method}}                                      & \textbf{Single-Hop} & \textbf{Multi-Hop} & \textbf{Temporal} & \textbf{Open-Domain} & \textbf{Adversarial} & \textbf{Overall} \\ \midrule
\textbf{Graph}                     & \textbf{QA Agent: Claude 4.5 Haiku w/ Tools} & \multicolumn{6}{c}{}                                                    \\ \cmidrule(r){1-2}
\multirow{3}{*}{APEX-MEM} & SchemaViewer, EntityLookUp          & 80.85\%    & 76.64\%   & 72.92\%  & 76.34\%     & 77.80\%     & 77.19\% \\
                          & +  GraphSQL                         & 80.78\%    & 79.75\%   & 82.29\%  & 78.00\%     & 81.16\%     & 79.45\% \\
                          & + Search                            & 85.46\%    & 84.74\%   & 79.17\%  & 89.18\%     & 87.22\%     & 87.00\%    \\ \bottomrule
\end{tabular}
}
\caption{APEX-MEM Ablations of different tools}
\label{tab:ablations-tools}
\end{table*}

\subsection{Results}

\textbf{RQ1: How does APEX-MEM compare to existing state-of-the-art for memory-based Tasks?} APEX-MEM achieves state-of-the-art performance across multiple conversational memory benchmarks, substantially outperforming previous systems. On the LOCOMO benchmark, APEX-MEM with GPT5 achieves 88.88\% overall accuracy (Table~\ref{tab:locomo_evals}), surpassing the previous best system MIRIX at 85.38\% by 3.50 percentage points. APEX-MEM demonstrates strong performance across all question categories: 89.88\% on single-hop, 86.29\% on multi-hop reasoning, 90.63\% on temporal queries, 91.68\% on open-domain, and 86.77\% on adversarial examples. With GPT4o as the QnA agent, APEX-MEM achieves 86.35\% overall, demonstrating that the architecture generalizes across LLM backends, though we observe higher error rates for GPT4o in SQLite query generation and tool usage. On the LongMemEval, APEX-MEM with Claude 4.5 Sonnet achieves 86.2\% overall score (Table~\ref{tab:longmem}), improving over the strongest baseline Nemori~\cite{nan2025nemori} at 74.6\% by 11.6 percentage points and session-aware RAG baselines at 72.5\% by 13.7 points. These results demonstrate that APEX-MEM provides a robust foundation for long-term conversational memory.

\begin{table*}[t]
\tiny
\resizebox{\linewidth}{!}{
\begin{tabular}{cccc}
\hline
\textbf{Method}                       &                                                    &                                       & \textbf{Overall Score }             \\ \hline
\multicolumn{1}{c}{APEX-MEM} & \multicolumn{1}{c}{\textbf{Graph}}                          & \multicolumn{1}{c}{\textbf{QA Agent}}          &                            \\ \cline{2-3}
\multicolumn{1}{c}{}         & \multicolumn{1}{c}{\multirow{5}{*}{APEX-MEM Online}}      & \multicolumn{1}{c}{Claude 4.5 Haiku}  & \multicolumn{1}{c}{82.8\%} \\
\multicolumn{1}{c}{}         & \multicolumn{1}{c}{}                               & \multicolumn{1}{c}{Claude 4.5 Sonnet} & \multicolumn{1}{c}{\textbf{86.2\%}}       \\
                             & \multicolumn{1}{c}{}                               & \multicolumn{1}{c}{Claude 4 Sonnet}   & 81.0\%                     \\
\multicolumn{1}{c}{}         & \multicolumn{1}{c}{}                               & \multicolumn{1}{c}{GPT5}              & \multicolumn{1}{c}{85.2\%} \\
                             & \multicolumn{1}{c}{}                               & \multicolumn{1}{c}{GPT4o}             & 75.0\%                     \\ \hline
\textbf{Baseline}                     & \textbf{Retrieval}                                          & \textbf{Agent}                                 &                            \\ \cline{2-3}
                             & Full-Context                                       & GPT4o                                 & 60.2\%                     \\
                             & Full-Context                                       & Claude 4.5 Sonnet                     & 62.2\%                     \\
                             & Full-Context + Chain-of-Note                       & Claude 4.5 Sonnet                     & 63.9\%                     \\
                             & SimpleSearch (K=V+fact), top-5 + expanded Sessions & Claude 4.5 Sonnet                     & 72.5\%                     \\
LongMemEval                  & SimpleSearch (K=V+fact)                            & Mistral-Nemo-Instruct-2407            & 66.6\%                     \\ \hline
\textbf{Memory}                       & \textbf{MemoryAgent}                                        &                                       &                            \\ \cline{2-2}
                             & Zep                                                &                                       & 71.2\%                     \\
                             & Mem0                                               &                                       & 71.3\%                    \\
                             & A-Mem                                              &                                       & 59.3\%                    \\
                             & Nemori                                             &                                       & 74.6\%                     \\ \hline
\end{tabular}}
\caption{LongMemEval Evaluation Results}
\label{tab:longmem}
\end{table*}

\textbf{RQ2: How important is each tool for APEX-MEM-based QnAAgent?} The ablation study in Table~\ref{tab:ablations-tools} reveals that each tool component contributes significantly to APEX-MEM's performance. Using only SchemaViewer and EntityLookUp tools, the system achieves 77.19\% overall accuracy on LOCOMO. Adding GraphSQL capabilities improves performance to 79.45\%, representing a 2.26\% gain, with particularly strong improvements on multi-hop reasoning (76.64\% to 79.75\%) and temporal queries (72.92\% to 82.29\%). The addition of the Search tool further boosts overall accuracy to 87\%, a 7.55 point improvement, with substantial gains across all categories including single-hop (80.78\% to 85.46\%), multi-hop (79.75\% to 84.74\%), open-domain (78.00\% to 89.18\%), and adversarial questions (81.16\% to 87.22\%). This demonstrates that the combination of entity linking, structured graph traversal via SQL, and hybrid semantic search is essential for achieving optimal performance in conversational memory tasks. In the appendix, Table~\ref{tab:tool_distribution} contains analysis on tool distribution for different ablation methods, and Figure~\ref{fig:tool-v/s-acc} for task v/s accuracy v/s \# of tool calls. GraphSQL and Search tools complement each other to enhance both quality and efficiency: while GraphSQL-only systems require 3.3x more tool calls (27,282 vs 8,260) to achieve 79.45\% accuracy, the hybrid approach leverages Search for rapid context retrieval and GraphSQL for precise structured reasoning, achieving superior 87\% accuracy with balanced tool usage. GraphSQL is a complex structured query generation task, as we are using SQLite to model graph-queries. In future we will evaluate the impact of Graph databases.

\begin{table}[t]
\centering
\small
\setlength{\tabcolsep}{4pt}
\begin{tabular}{@{}llr@{}}
\toprule
\multicolumn{2}{l}{\textbf{Method}} & \textbf{Acc.} \\ \midrule
\multicolumn{1}{l|}{\multirow{5}{*}{\shortstack[l]{APEX-MEM\\(Online)}}} & QnAAgent              &          \\
\multicolumn{1}{l|}{}                          & Claude 4 Sonnet       & 28.9\%   \\
\multicolumn{1}{l|}{}                          & Claude 4.5 Sonnet     & 35.2\%   \\
\multicolumn{1}{l|}{}                          & GPT5                  & \textbf{40.1}\% \\
\multicolumn{1}{l|}{}                          & GPT4o                 & 19.0\%   \\ \midrule
\multicolumn{2}{l}{\textbf{Baselines w/ Web-Search Tool}} &          \\ \midrule
\multicolumn{1}{l|}{\multirow{6}{*}{AGENT}}    & GPT4o                 & 15\%     \\
\multicolumn{1}{l|}{}                          & GPT5                  & 38.6\%   \\
\multicolumn{1}{l|}{}                          & O4-Mini-HIGH          & 12\%     \\
\multicolumn{1}{l|}{}                          & O3                    & 34.6\%   \\
\multicolumn{1}{l|}{}                          & QWEN3-235B            & 11.4\%   \\
\multicolumn{1}{l|}{}                          & DeepSeek-R1           & 15.4\%   \\ \bottomrule
\end{tabular}
\caption{SealQA Evaluation Results}
\label{tab:sealqa}
\end{table}

\textbf{RQ3: Does APEX-MEM improve QA Agent's ability to handle complex scenarios compared to other Deep research agents?} APEX-MEM demonstrates superior performance on complex, multi-document reasoning tasks compared to state-of-the-art research agents. On the SealQA-Hard benchmark (Table~\ref{tab:sealqa}), which evaluates search-augmented LLMs on challenging fact-seeking questions with conflicting and noisy web search results, APEX-MEM with GPT5 achieves 40.15\% accuracy, substantially outperforming baseline agents including O3 at 34.6\%, DeepSeek-R1 at 15.4\%, GPT4o at 15\%, and O4-Mini-HIGH at 12\%. This 5.55 percentage point improvement over the strongest baseline demonstrates APEX-MEM's ability to effectively resolve contradictions and filter noise through its append-only property graph architecture and retrieval-time resolution strategy. The system's ontology-grounded entity linking and GraphSQL traversal enable more precise identification of relevant information across multiple conflicting sources, a critical capability for real-world applications where information quality varies significantly. See Table \ref{tab:sql_examples} for GraphSQL examples.

\textbf{RQ4: How well does APEX-MEM generalize?} APEX-MEM generalizes across diverse reasoning types and task categories. On LOCOMO (Table~\ref{tab:locomo_evals}), the system maintains consistently high performance across different question types with less than 5 percentage points variation. This contrasts sharply with previous systems such as MIRIX (20\% drop in temporal accuracy 65.62\% temporal) and Mem0 (10\% drop in single-hop and 18\% drop in multi-hop), which show significant performance disparities across question types. Furthermore, APEX-MEM's strong results on both LOCOMO (88.88\%), LongMemEval (86.2\%), and SealQA-Hard (40.15\%) demonstrate effective generalization across different benchmarks, conversation lengths, and information quality conditions, validating the domain-agnostic nature of the ontology-supported property graph architecture.

%MIRIX (85.11\% single-hop, 83.70\% multi-hop, 65.62\% temporal) and Mem0 (65.71\% single-hop, 47.19\% multi-hop, 75.71\% temporal)

% \subsection{LongMemEval}

% \subsection{SealQA}

\section{Conclusion}

We have introduced APEX-MEM, an ontology-supported property graph architecture for long-term conversational memory that achieves state-of-the-art performance across multiple challenging benchmarks. By combining append-only semantics at construction time with retrieval-time resolution through a graph agent equipped with entity linking, GraphSQL traversal, and hybrid search capabilities, APEX-MEM addresses fundamental limitations in existing memory systems. Our approach achieves 88.88\% accuracy on LOCOMO's Question Answering task, surpassing the previous best system MIRIX by 3.50 percentage points, and 86.2\% on LongMemEval, improving over strong session-aware RAG baselines by 13.7 points. These results demonstrate that property graphs provide a robust foundation for capturing rich semantic information while enabling efficient retrieval and intelligent conflict resolution in conversational contexts.

Our evaluation focuses on conversational Question Answering tasks across LOCOMO-QA, LongMemEval, and SealQA-Hard. APEX-MEM is architecturally designed as a retrieval-augmented QA system optimized for query-driven fact selection and temporal resolution. Extending to Event Summarization and Multimodal Dialog Generation---which require generative narrative synthesis and different evaluation metrics---represents valuable future work that builds on our append-only temporal memory foundation.

Despite these advances, significant opportunities remain for future work. First, while APEX-MEM substantially outperforms existing systems, performance gaps persist in SealQA-Hard (40.15\% accuracy), where noisy, conflicting multi-document scenarios continue to challenge even our best models. Improving fact extraction to represent more complete entity information—including better handling of implicit relationships, temporal nuances, and contextual dependencies—will be critical for closing this gap. Second, our current graph agent requires multiple tool invocations to converge on solutions, with the ablation study showing that all three tool types (EntityLookUp, GraphSQL, and Search) contribute meaningfully to performance. Future work will focus on developing more efficient query planning strategies and learned retrieval policies that reduce the number of agent tool calls while maintaining or improving answer quality. Third, extending the ontology to capture domain-specific knowledge structures and exploring automated ontology refinement from conversational data could further enhance the system's ability to represent and reason over specialized knowledge domains. Finally, investigating the integration of APEX-MEM with emerging long-context models and exploring hybrid architectures that combine parametric memory with our structured external memory approach represent promising directions for achieving human-level performance across all conversational memory tasks.

\section*{Limitations}
While APEX-MEM demonstrates strong performance across multiple benchmarks and reasoning tasks, several limitations remain and highlight areas for future exploration.

First, the graph construction process incurs significant computational costs, particularly during entity resolution and property extraction phases. The current implementation relies on large language models for fact extraction, which can be resource-intensive for real-time conversational applications. Future work should explore smaller, more efficient models and optimized algorithms to reduce construction overhead while maintaining graph quality and precision. Additionally, the accuracy of the constructed graph depends heavily on the quality of entity and property resolution—errors or ambiguities in these processes can propagate through the graph structure, potentially affecting downstream retrieval and reasoning performance.

Second, while APEX-MEM employs a domain-agnostic ontology with 35 entity classes, standardizing ontology schemas across different conversational domains remains challenging. The current ontology may not capture all domain-specific nuances, and queries requiring highly specialized knowledge structures may benefit from extended or customized ontologies. Developing standardized ontology frameworks and exploring automated ontology refinement from conversational data could enhance the system's ability to support more precise querying and reasoning across diverse application domains.

Third, APEX-MEM's performance is sensitive to the QnA agent's ability to generate correct SQLite queries and use tools effectively. With GPT4o as the QnA agent, we observed critically high error rates in graph query generation and tool selection, resulting in 86.35\% overall accuracy on LOCOMO compared to 88.88\% with GPT5. We had to add explicit query generation error examples to the GPT4o prompt to achieve this level. Both Claude Sonnet 3.5 and 4.5 show similar tool-use success rates and comparable performance, suggesting that the architecture works best with models that natively support structured tool interactions. This dependency on the base model's tool-use capabilities represents a practical deployment consideration.

Additionally, APEX-MEM's performance on particularly SealQA-Hard (40.15\% accuracy) indicates room for improvement in handling noisy, conflicting multi-document scenarios. The current graph agent also requires multiple tool invocations to converge on solutions, which can impact response latency in interactive settings. Optimizing the agent's reasoning strategy and reducing the number of tool calls needed would improve efficiency without sacrificing accuracy.

Finally, we acknowledge that APEX-MEM's current implementation is limited to text-based interactions. Future work could extend the system to support multimodal inputs, such as images, audio, or video, enabling richer and more comprehensive contextual representations. Integration with emerging long-context models and exploration of hybrid architectures that combine parametric memory with structured external memory also present promising directions for enhancing the system's capabilities.

\section*{Ethical Considerations}

We have conducted a comprehensive review of all scientific artifacts utilized in this research, including datasets and computational models, to ensure their licenses explicitly permit academic research and publication use. All datasets employed in our experiments have been properly de-identified to maintain participant anonymity and protect individual privacy.

Our proposed APEX-MEM framework offers significant potential for reducing both the economic and environmental impact associated with LLM enhancement. By minimizing the requirements for extensive data collection and manual annotation processes for Knowledge Graph construction, our approach streamlines model development while providing robust safeguards for user privacy and data protection. This reduction in data collection needs helps mitigate the risk of information leakage during training corpus assembly and reduces the computational resources typically required for model improvement. Moreover, we aim to further reduce LLM inference costs by improving smaller LLMs by using Knowledge Graphs.

Throughout the development and evaluation of this work, generative AI systems were employed in a limited and transparent manner. Specifically, AI assistance was utilized for language refinement, paraphrasing, and grammatical checking during the manuscript preparation process. Additionally, generative AI was used to assist in writing code test cases for system validation and served as an automated judge for evaluation tasks, supplementing human evaluation where appropriate. All core research contributions, experimental design, analysis, and conclusions presented in this paper are the result of human intellectual effort and scientific reasoning.

% \section*{Acknowledgments}

% Bibliography entries for the entire Anthology, followed by custom entries
%\bibliography{anthology,custom}
% Custom bibliography entries only
\bibliography{custom}

\appendix

\section{Appendix}
\label{sec:appendix}

% Tool Call Distribution Table
\begin{table*}[htbp]

\centering
\resizebox{\linewidth}{!}{
\begin{tabular}{lrrrr}
\toprule
\textbf{Tool} & \textbf{Claude 4.5 Haiku} & \textbf{Claude 4.5 Haiku: Entity} & \textbf{Claude 4.5 Haiku: \textsc{GraphSQL}} & \textbf{Claude 4.5 Sonnet} \\
\midrule
\textsc{EntityLookup} & 5160 & 17414 & 5546 & 4894 \\
\textsc{Search} & 8900 & 1058 & 0 & 10544 \\
\textsc{PropertySearch} & 3346 & 358 & 5016 & 3342 \\
\textsc{SchemaViewer} & 3958 & 4004 & 3622 & 3964 \\
\textsc{GraphSQL} & 8260 & 0 & 27282 & 7168 \\
\bottomrule
\end{tabular}}
\caption{Tool Call Distribution Across Methods}
\label{tab:tool_distribution}
\end{table*}

\subsection{Comparative Analysis: APEX-MEM vs. GraphSQL-Only Retrieval}

To demonstrate the value of APEX-MEM's hybrid tool architecture, we compare the full system against a GraphSQL-only ablation that relies exclusively on SQL queries for memory retrieval. This analysis reveals how APEX-MEM's multi-tool approach achieves superior performance through complementary retrieval strategies.

\subsubsection{Tool Usage Patterns and Strategic Differences}

Table~\ref{tab:tool_distribution} illustrates the contrasting tool usage patterns between APEX-MEM and its GraphSQL-only variant. APEX-MEM employs a balanced retrieval strategy with 8,260 GraphSQL calls, 8,900 Search calls, and 5,160 EntityLookup calls, demonstrating effective utilization of all available tools. In contrast, the GraphSQL-only agent compensates for the absence of hybrid retrieval by invoking GraphSQL 27,282 times—a 3.3× increase over the full system. This dramatic escalation reflects the agent's need to express all reasoning operations through declarative queries, even when alternative retrieval methods might be more efficient.

Notably, while the GraphSQL-only variant increases PropertySearch calls from 3,346 to 5,016 and EntityLookup calls from 5,160 to 5,546. These auxiliary tools remain essential for identifying entity IDs and property names required to construct valid SQL queries, highlighting that even SQL-centric approaches depend on entity resolution capabilities.

\subsubsection{Performance Implications}

The architectural differences translate directly into performance outcomes. APEX-MEM achieves 87.00\% overall accuracy on LOCOMO, outperforming the GraphSQL-only ablation (79.45\%) by 7.55 percentage points. This advantage is particularly pronounced in open-domain questions, where APEX-MEM's 89.18\% accuracy surpasses the GraphSQL variant by 11.18 points, demonstrating the critical value of the Search tool for retrieving information not easily accessible through structured queries alone.

Interestingly, the GraphSQL-only agent achieves competitive or superior performance on temporal reasoning tasks (82.29\% vs. 79.17\%), where SQL's native temporal operators and date functions provide natural expressiveness. Table~\ref{tab:sql_categories} reveals that temporal queries constitute the largest category (6,775 instances) in the GraphSQL-only configuration, followed by aggregate queries (4,586), SELECT queries (1,850), and JOIN operations (430). These statistics underscore SQL's strength in temporal and quantitative reasoning, while highlighting the limitations of a purely query-based approach for diverse conversational memory tasks.

% SQL Query Categories Table
\begin{table*}[]
\centering
\resizebox{\linewidth}{!}{
\begin{tabular}{lrrrr}
\toprule
SQL Category & Claude 4.5 Haiku & Claude 4.5 Haiku: Entity & Claude 4.5 Haiku: GraphSQL & Claude 4.5 Sonnet \\
\midrule
SELECT & 315 & 0 & 1850 & 88 \\
JOIN & 81 & 0 & 430 & 31 \\
AGGREGATE & 1417 & 0 & 4586 & 1230 \\
TEMPORAL & 2317 & 0 & 6775 & 2235 \\
% OTHER & 0 & 0 & 0 & 0 \\
\bottomrule
\end{tabular}}
\caption{SQL Query Categories by Method}
\label{tab:sql_categories}
\end{table*}

\subsubsection{Query Complexity and Reasoning Patterns}

Table~\ref{tab:sql_examples} provides representative examples of the SQL queries generated by both systems. The temporal query example illustrates the sophistication achievable through declarative queries, computing multiple temporal metrics (days, weeks, approximate months) using SQLite's Julian day functions. The aggregate query demonstrates set operations over distinct entities, while the JOIN example shows relationship traversal across facts and evidence tables.

However, it is important to note that these query categories are not mutually exclusive—a single conversational question often requires multiple SQL queries of different types to construct a complete answer. APEX-MEM mitigates this complexity by strategically selecting the most appropriate tool for each reasoning step, reducing the total number of tool invocations needed to converge on correct answers. The GraphSQL-only variant, lacking this flexibility, must decompose complex questions into multiple SQL operations, contributing to both increased query volume and reduced overall accuracy.

\subsubsection{Implications for System Design}

This comparative analysis demonstrates that while GraphSQL provides powerful capabilities for structured reasoning—particularly for temporal and quantitative queries—APEX-MEM's hybrid architecture achieves superior performance by leveraging complementary tool strengths. The full system balances structured queries with entity-based retrieval and semantic search, enabling more efficient and accurate conversational memory access across diverse reasoning requirements.

% SQL Query Examples Table
\begin{table*}[]
\centering
\resizebox{\linewidth}{!}{
\begin{tabular}{p{2cm}p{12cm}}
\toprule
Category & Example Query \\
\midrule
SELECT & \texttt{SELECT DISTINCT entity\_id, entity\_name, entity\_type FROM entities WHERE entity\_name LIKE '\%Anthony\%' COLLATE NOCASE} \\
JOIN & \texttt{SELECT f.property\_name, f.value\_json, f.dtype FROM facts f JOIN evidence e ON e.fact\_id = f.id WHERE e.event\_id = 518} \\
AGGREGATE & \texttt{SELECT COUNT(DISTINCT device\_name) as device\_count FROM ( SELECT 'Fitbit Versa 3' as device\_name UNION SELECT 'nebulizer machine' as device\_name UNION SELECT 'Accu-Chek Aviva Nano' as device\_name UNION SELECT 'hearing aids' as device\_name )} \\
TEMPORAL & \texttt{SELECT f.value\_json as start\_date, date(:question\_date) as question\_date, julianday(:question\_date) - julianday(json\_extract(f.value\_json, '\$')) as days\_difference, CAST((julianday(:question\_date) - julianday(json\_extract(f.value\_json, '\$')))/ 30.44 AS INTEGER) as months\_approx, CAST((julianday(:question\_date) - julianday(json\_extract(f.value\_json, '\$')))/7 AS INTEGER) as weeks\_approx FROM facts f WHERE f.subject\_id = :user\_id AND f.property\_name = :property\_id ORDER BY f.created\_at DESC LIMIT 1}\\
\bottomrule
\end{tabular}}
\caption{Sample SQL Queries by Category}
\label{tab:sql_examples}
\end{table*}

\begin{figure*}[]
  \centering
 \includegraphics[width=\textwidth]{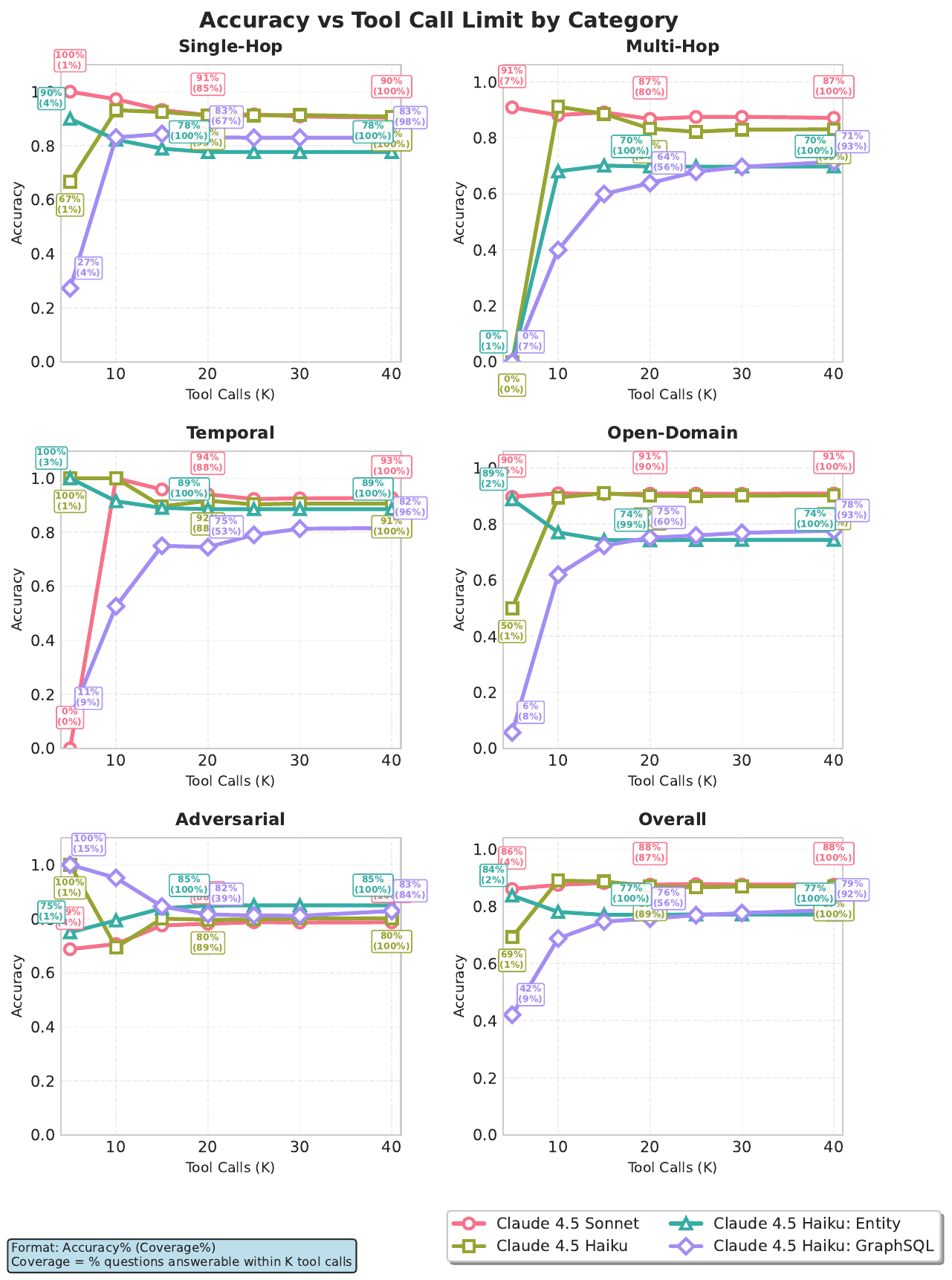}
\caption{Analysis of Tool Calls v/s Accuracy on LOCOMO Dataset. We cap the max Tool calls at 40.}
\label{fig:tool-v/s-acc}
\end{figure*}

\section{Tool Call Efficiency Analysis on LOCOMO}

Figure~\ref{fig:tool-v/s-acc} presents a comprehensive analysis of the relationship between tool call frequency and accuracy across different question categories on the LOCOMO dataset, with tool calls capped at 40. This analysis reveals distinct behavioral patterns depending on the agent's available tool set and the underlying language model's capabilities.

\subsection{Agent-Specific Behavioral Patterns}

The tool call distribution demonstrates that different agent configurations exhibit markedly different behaviors as the number of tool calls increases. APEX-MEM with Claude 4.5 Sonnet achieves the highest efficiency, reaching approximately 84-86\% accuracy with just 10 tool calls across most categories and answering 80-90\% of questions within the first 10-20 tool calls. 
 The full APEX-MEM system with Claude 4.5 Haiku shows similar patterns but requires slightly more tool calls to converge to comparable accuracy levels.

In contrast, the GraphSQL-only ablation requires significantly more tool calls to achieve competitive performance. This agent must first discover the graph structure through SchemaViewer and EntityLookup operations before constructing effective SQL queries, resulting in a more gradual accuracy improvement curve. However, with sufficient tool calls (approximately 20-30), the GraphSQL-only agent outperforms the EntityLookup-only baseline across most categories, demonstrating that structured queries can eventually compensate for the absence of hybrid retrieval capabilities. 

\subsection{Adversarial Questions: The Noise Introduction Effect}

Adversarial questions exhibit a unique pattern where increased tool calls can introduce noise rather than improve accuracy. At 10 tool calls, Claude 4.5 Sonnet achieves 100\% accuracy (albeit with only 1\% coverage), while the full Haiku system reaches 75\% accuracy. 
 However, as tool calls increase to 20, all agents converge to relatively stable performance levels: Sonnet at 85\%, Haiku at 80\%, and GraphSQL at 78\% accuracy. 
 Beyond 20 tool calls, performance remains largely constant, with all methods eventually converging to 80-85\% accuracy at 40 tool calls. 

This convergence pattern suggests that for adversarial questions—which are specifically designed to be misleading or contain conflicting information—additional tool invocations may retrieve contradictory evidence that introduces uncertainty into the reasoning process. Most agents effectively reach their performance ceiling at approximately 20 tool calls, after which further retrieval provides diminishing or even slightly negative returns.

\subsection{Consistency of Stronger LLM Agents}

For non-adversarial categories (single-hop, multi-hop, temporal, and open-domain questions), stronger LLM agents demonstrate remarkable consistency in accuracy regardless of tool call count. APEX-MEM with Claude 4.5 Sonnet maintains high accuracy even with relatively few tool calls, requiring at least 10 tool calls to confidently answer approximately 80\% of queries. 
 This efficiency reflects the model's superior reasoning capabilities and its ability to strategically select the most informative tool invocations.

The full APEX-MEM system reaches 87-91\% accuracy across most categories with 40 tool calls, while the GraphSQL-only variant plateaus at 79-83\% depending on the question type. 
 Notably, the GraphSQL agent shows competitive performance on temporal queries (82\% accuracy), where SQL's native temporal operators provide natural expressiveness, but struggles more significantly on open-domain questions (78\% maximum), where the absence of the Search tool limits its ability to retrieve relevant contextual information. 

\subsection{Graph Discovery and Multi-Call Requirements}

The GraphSQL-only agent's performance trajectory illustrates the inherent cost of graph discovery in structured query-based approaches. Unlike the full APEX-MEM system, which can leverage EntityLookup and Search tools for rapid information access, the GraphSQL agent must iteratively explore the graph schema, identify relevant entities and properties, and construct appropriate SQL queries. This discovery process manifests as a more gradual accuracy improvement curve, with the agent requiring 20-30 tool calls to reach performance levels that APEX-MEM achieves with 10-15 calls. 

Despite this efficiency gap, the GraphSQL-only agent's ability to eventually outperform the EntityLookup-only baseline (which plateaus at approximately 77\% overall accuracy) demonstrates the value of structured querying for complex reasoning tasks. 
 The EntityLookup-only approach, while efficient for simple entity retrieval, lacks the compositional reasoning capabilities needed for multi-hop, temporal, and aggregate queries, resulting in consistently lower maximum accuracy across all categories.

These findings underscore the importance of APEX-MEM's hybrid tool architecture, which balances the efficiency of direct entity lookup, the expressiveness of structured queries, and the flexibility of semantic search to achieve superior performance with minimal tool call overhead.

\subsection {Additional Future Work}
While APEX-MEM achieves state-of-the-art accuracy, the current implementation faces efficiency challenges that warrant future investigation. Our analysis reveals that the system requires multiple tool invocations to converge on solutions, with most agents reaching their performance ceiling at approximately 20 tool calls. The 20-call performance ceiling and diminishing returns beyond this threshold suggest opportunities for intelligent stopping criteria.

\noindent\textbf{improve query planning}: 1. Develop reinforcement learning-based query planning that: 1/Predicts optimal tool sequences based on question type, 2/ Learns when to stop retrieval, 3/ Reduces average tool calls from 20-30 to 10-15 while maintaining accuracy
\noindent\textbf{efficient Model Alternatives}: Claude 4.5 Sonnet's superior efficiency (84-86\% at 10 calls) demonstrates that stronger reasoning reduces tool call requirements.
\noindent\textbf{improve efficiency in graph construction}: 1/ Fine-tune smaller models (7B-14B parameters) for specific subtasks to achieve Sonnet-level efficiency at lower cost, 2/ Target the 95-97\% fact extraction precision and 95-98\% entity resolution accuracy, and 3/ Explore specialized models for entity resolution, property extraction, and query planning
3.GraphSQL-only requires 3.3× more calls due to schema discovery overhead, while EntityLookup-only plateaus at 77\%.
\noindent\textbf{improve tool usage efficiency}: 1/ Investigate adaptive tool selection strategies that choose optimal tools based on query characteristics, 2/ Develop caching mechanisms for frequently accessed schema patterns to reduce discovery overheads, 3/ Explore parallel tool execution when dependencies allow.

\section{Cost and Resource Comparison}
\label{sec:cost}

Table~\ref{tab:tokencost} presents a comprehensive token-based cost comparison across all memory methods. Token counts provide a pricing-agnostic, stable comparison metric, as dollar costs vary by model and change frequently. Every memory method has two cost components: (1) \textbf{Graph/Memory Construction (GC)}---a one-time cost per conversation, amortized over queries, and (2) \textbf{Query Answering (QnA)}---per-query retrieval and generation cost. Graph construction accounts for only 16.6\% of APEX-MEM's total cost; the majority is spent on tool access and agentic reasoning loops. Overall, APEX-MEM's per-query token consumption is significantly lower than MIRIX, which performs approximately 8 LLM calls per interaction.

\begin{table*}[t]
\centering
\resizebox{\linewidth}{!}{
\begin{tabular}{lccccccc}
\toprule
\textbf{Method} & \textbf{GC} & \textbf{GC} & \textbf{Amort.} & \textbf{Mem} & \textbf{QnA} & \textbf{Total} & \textbf{Acc} \\
                & \textbf{Tok/Conv} & \textbf{Calls} & \textbf{GC/Q} & \textbf{Tok/Q} & \textbf{Tok/Q} & \textbf{Tok/Q} & \textbf{(w/o Adv)} \\
\midrule
MIRIX (est.) & $\sim$15.2M & $\sim$4,704 & $\sim$98,750 & $\sim$4,500 & $\sim$13,500 & $\sim$112,000 & 85.38\% \\
Zep/Graphiti & $\sim$9.4M & $\sim$5,292 & $\sim$60,900 & 2,247 & $\sim$3,900 & $\sim$64,800 & 75.14\% \\
Mem0$^g$ (est.) & $\sim$4.9M & $\sim$2,352 & $\sim$31,882 & 3,616 & $\sim$5,300 & $\sim$37,200 & 68.44\% \\
\textbf{APEX-MEM} & 2.69M & 3,717 & 13,557 & 8,000$^\ddagger$ & $\sim$16,000 & $\sim$30,000 & 84--89\%$^\dagger$ \\
Full Context & 0 & 0 & 0 & 23,653 & $\sim$25,000 & $\sim$25,000 & 87.52\% \\
Mem0 (est.) & $\sim$1.9M & $\sim$882 & $\sim$12,409 & 1,764 & $\sim$3,500 & $\sim$15,900 & 68.44\% \\
Nemori (est.) & $\sim$1.0M & $\sim$765 & $\sim$6,422 & 2,745 & $\sim$4,500 & $\sim$10,900 & 79.40\% \\
\bottomrule
\end{tabular}
}
\caption{Token consumption comparison across memory methods. $^\dagger$Accuracy depends on QnA model. $^\ddagger$Tunable via top-$k$ retrieval budget. GC = Graph/Memory Construction; Amort. = Amortized over queries per conversation.}
\label{tab:tokencost}
\end{table*}

Table~\ref{tab:tokendecomp} decomposes APEX-MEM's per-query token usage by component, revealing that memory retrieval content (26.6\%) and tool framing overhead (27.3\%) dominate, while graph construction is amortized to only 16.6\%.

\begin{table}[htbp]
\centering
\footnotesize
\setlength{\tabcolsep}{3pt}
\begin{tabular}{lccp{2.5cm}}
\toprule
\textbf{Component} & \textbf{Tok/Q} & \textbf{\%} & \textbf{Source} \\
\midrule
Graph constr. (amort.) & 13,557 & 16.6 & Extraction, resolution \\
System prompt & 7,854 & 9.6 & Fixed per architecture \\
Memory retrieval & 21,745 & 26.6 & Graph search (top-$k$) \\
Agent loop overhead & 16,174 & 19.8 & Prior msgs in loop \\
Tool framing & 22,274 & 27.3 & Tool call/response fmt \\
\midrule
\textbf{Total (mean)} & \textbf{81,604} & \textbf{100} & \\
\bottomrule
\end{tabular}
\caption{APEX-MEM token decomposition per query.}
\label{tab:tokendecomp}
\end{table}

\section{GraphSQL Execution Statistics}
\label{sec:graphsql_stats}

Table~\ref{tab:graphsql_exec} presents GraphSQL execution analysis across different LLM backends. Claude Sonnet 4.5 achieves the highest success rate (97.6\%), while GPT-5 shows a higher error rate (6.6\%), primarily due to SQLite syntax differences.

\begin{table}[htbp]
\centering
\footnotesize
\setlength{\tabcolsep}{2pt}
\begin{tabular}{lcccc}
\toprule
\textbf{Metric} & \textbf{Sonnet} & \textbf{GPT-5} & \textbf{Sonnet} & \textbf{Haiku} \\
                 & \textbf{4.5}    &                & \textbf{(SQL)}  &                \\
\midrule
SQL Executions & 3,659 & 2,163 & 66,580 & 4,277 \\
Successful & 3,574 & 2,020 & 65,900 & 4,080 \\
SQL Errors & 87 & 143 & 791 & 197 \\
\textbf{Success Rate} & \textbf{97.6\%} & 93.4\% & \textbf{98.8\%} & 95.4\% \\
\bottomrule
\end{tabular}
\caption{GraphSQL execution analysis across LLM backends. ``Sonnet (SQL)'' denotes the GraphSQL-only ablation.}
\label{tab:graphsql_exec}
\end{table}

The agent successfully recovered from 87\% of SQL failures through three mechanisms: (1) \textbf{SchemaViewer consultation} (45\% of recoveries): re-examining table structure and correcting syntax; (2) \textbf{Fallback to EntityLookup} (28\%): retrieving structured entity documents when SQL is too complex, then identifying correct properties for structured queries; (3) \textbf{Fallback to Search} (14\%): using semantic retrieval when graph structure is insufficient. This demonstrates the robustness of the multi-tool architecture: no single tool failure blocks progress, and the agent adapts its retrieval strategy based on tool success.

\section{Qualitative Case Studies}
\label{sec:case_studies}

We present three case studies illustrating APEX-MEM's temporal resolution, multi-hop reasoning, and failure modes.

\paragraph{Case 1: Temporal Contradiction Resolution.}
\textit{Question:} ``What is Alice's current favorite restaurant?''

\noindent\textit{Timeline:} Session~1 (2024-01-15): Alice says ``I love Italian Garden! Their pasta is the best in town.'' Session~5 (2024-03-20): Alice says ``Italian Garden closed down last month. Now I go to Sakura Sushi every week instead.''

\noindent\textit{APEX-MEM Processing:} During construction, the system extracts \texttt{(Alice, favorite\_restaurant, ``Italian Garden'', from=2024-01-15)} and later \texttt{(Alice, favorite\_restaurant, ``Sakura Sushi'', from=2024-03-20)} \textbf{without deleting the previous fact}. At retrieval time, a GraphSQL temporal query returns both facts ordered by timestamp; the agent selects the most recent valid entry (``Sakura Sushi''). This preserves the full temporal evolution with evidence links, enabling follow-up questions like ``When did Alice's favorite restaurant change?''

\paragraph{Case 2: Multi-Hop Reasoning Success.}
\textit{Question:} ``What is the title of Bob's manager?''

\noindent\textit{Processing:} The agent performs multi-hop traversal via GraphSQL: Bob $\rightarrow$ \texttt{reports\_to} $\rightarrow$ Sarah Chen $\rightarrow$ \texttt{job\_title} $\rightarrow$ VP of Engineering. The structured graph enables compositional queries that would be difficult with flat text retrieval.

\paragraph{Case 3: Failure Mode---Entity Linking Error.}
\textit{Question:} ``How many times did Bob visit restaurants in Paris last month?''

\noindent\textit{Root Cause:} Entity linking failed to connect restaurant names (``Le Jules Verne'') to the Paris location entity from contextual clues (``Eiffel Tower''). The system correctly identified the restaurants as entities but did not resolve the implicit spatial relationship.

\noindent\textit{Lesson:} High-quality entity linking is critical for spatial/relational queries. Integrating external knowledge bases (e.g., Wikidata) for landmark-to-location resolution could address this limitation.

\section{Append-Only vs.\ Eager Update Strategies}
\label{sec:appendonly}

Table~\ref{tab:appendonly} provides indirect evidence supporting APEX-MEM's append-only design by comparing temporal reasoning accuracy across systems with different update strategies.

\begin{table}[htbp]
\centering
\footnotesize
\setlength{\tabcolsep}{3pt}
\begin{tabular}{lp{1.8cm}cc}
\toprule
\textbf{System} & \textbf{Append-Only?} & \textbf{Temporal} & \textbf{$\Delta$} \\
\midrule
\textbf{APEX-MEM} & Yes & \textbf{90.63\%} & --- \\
Mem0 & No (consolid.) & 75.71\% & --14.92 \\
MIRIX & No (state merge) & 65.62\% & --25.01 \\
Zep & Partial (temp. KG) & 76.60\% & --14.03 \\
\bottomrule
\end{tabular}
\caption{LOCOMO temporal accuracy: append-only vs.\ eager update strategies. $\Delta$ is the difference from APEX-MEM in percentage points.}
\label{tab:appendonly}
\end{table}

APEX-MEM's +14 to +25 point advantage on temporal queries strongly suggests that preserving full history enables better temporal reasoning. Systems that consolidate or overwrite facts during construction lose the temporal provenance needed to resolve ``when did X change?'' or ``what was Y before Z?'' queries, which require access to superseded facts.

\section{APEX-MEM Construction and Evolution}
\label{sec:text_to_graph}

Table~\ref{tab:text_to_graph_conversion} demonstrates how APEX-MEM transforms conversational text into structured graph representations. The examples show the systematic extraction of entities, properties, and typed values from natural language, illustrating the rich semantic structure captured by our ontology-guided approach.

\begin{table*}[t]
\small
\begin{tabular}{@{}p{2.8cm}|p{2.2cm}|p{1.8cm}|p{2.5cm}|p{3.2cm}|p{1.2cm}@{}}
\toprule
\textbf{Original Text} & \textbf{Entities} & \textbf{Types} & \textbf{Properties} & \textbf{Values} & \textbf{Data Types} \\
\midrule

\multirow{4}{2.8cm}{"I went to a LGBTQ support group yesterday and it was so powerful."} 
& Caroline & PERSON & P:attended\_event & "LGBTQ support group" & str \\
\cline{2-6}
& & & P:experience\_ description & "powerful" & str \\
\cline{2-6}
& LGBTQ support group & EVENT & P:event\_date & "2023-05-07" & date \\
\cline{2-6}
& & & P:event\_type & "support group" & str \\
\midrule

\multirow{6}{2.8cm}{"I just signed up for a pottery class yesterday. It's like therapy for me."} 
& Melanie & PERSON & P:activities\_ participated & "pottery class" & str \\
\cline{2-6}
& & & P:activity\_benefit & "therapy for me" & str \\
\cline{2-6}
& & & P:emotional\_outlet & "pottery helps express emotions" & str \\
\cline{2-6}
& pottery class & EVENT & P:enrollment\_date & "yesterday" & date \\
\cline{2-6}
& pottery & CREATIVE\_ WORK & P:purpose & "self-expression" & str \\
\cline{2-6}
& & & P:therapeutic\_value & true & bool \\
\midrule

\multirow{4}{2.8cm}{"You'd be a great counselor! Your empathy will help people."} 
& Melanie & PERSON & P:career\_suggestion & "counselor" & str \\
\cline{2-6}
& & & P:personal\_qualities & "empathy" & str \\
\cline{2-6}
& & & P:predicted\_impact & "will help people" & str \\
\cline{2-6}
& counselor & PERSON & P:required\_skills & "empathy" & str \\
\midrule

\multirow{4}{2.8cm}{"I'm swamped with the kids \& work."} 
& Melanie & PERSON & P:has\_children & true & bool \\
\cline{2-6}
& & & P:employment\_status & "working" & str \\
\cline{2-6}
& & & P:emotional\_state & "overwhelmed" & enum \\
\cline{2-6}
& & & P:relationship\_status & "swamped with kids and work" & str \\
\midrule

\multirow{5}{2.8cm}{"The transgender stories were so inspiring! I was so thankful."} 
& Caroline & PERSON & P:emotional\_state & "thankful" & str \\
\cline{2-6}
& & & P:reaction\_to\_content & "inspiring" & str \\
\cline{2-6}
& transgender stories & CREATIVE\_ WORK & P:content\_type & "personal narratives" & str \\
\cline{2-6}
& & & P:emotional\_impact & "inspiring" & str \\
\cline{2-6}
& support & TOPIC & P:context & "LGBTQ community" & str \\

\bottomrule
\end{tabular}
\caption{Examples of text-to-graph conversion in APEX-MEM. Each conversational utterance is systematically transformed into structured entity-property-value triplets with appropriate semantic types, enabling sophisticated querying and temporal reasoning over conversational knowledge.}
\label{tab:text_to_graph_conversion}
\end{table*}

\section{Graph Structure Visualization}
\label{sec:graph_structure}

\begin{figure*}[t]
  \centering
 \includegraphics[width=\textwidth]{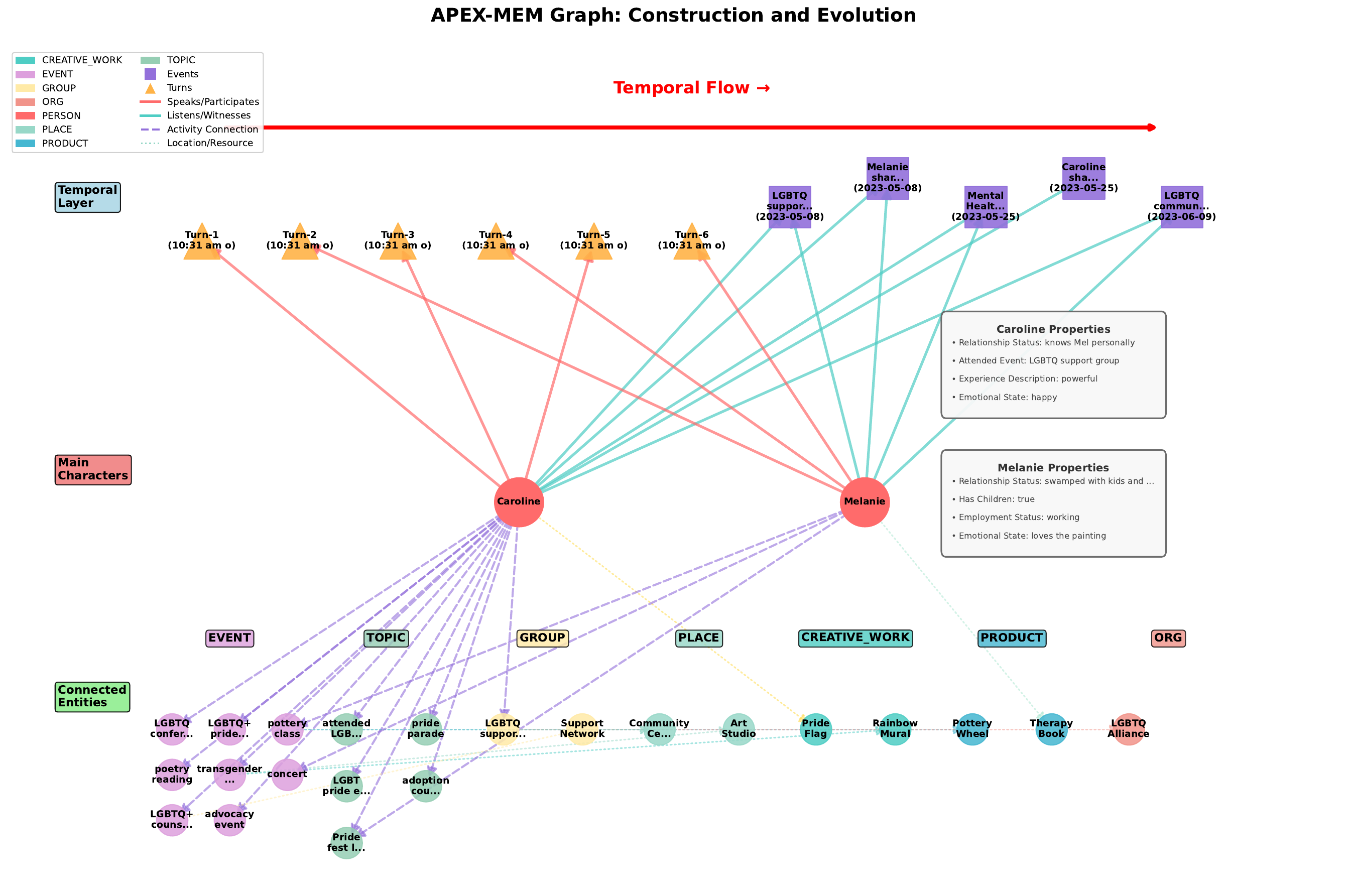}
\caption{APEX-MEM Graph Structure: The figure demonstrates how conversational turns and events connect to entities through participation relationships, with temporal information preserved and entities organized by semantic type for efficient querying and reasoning.}
\label{fig:apex_mem_graph}
\end{figure*}

Figure~\ref{fig:apex_mem_graph} illustrates the layered structure of APEX-MEM's property graph, showing how conversational elements are organized into temporal, entity, and relationship layers. The visualization demonstrates the systematic organization of turns, events, entities, and their properties in a coherent graph structure that enables efficient querying and reasoning.

\section{Ontological Architecture}
\label{sec:ontology}

\begin{figure*}[t]
  \centering
 \includegraphics[width=\textwidth]{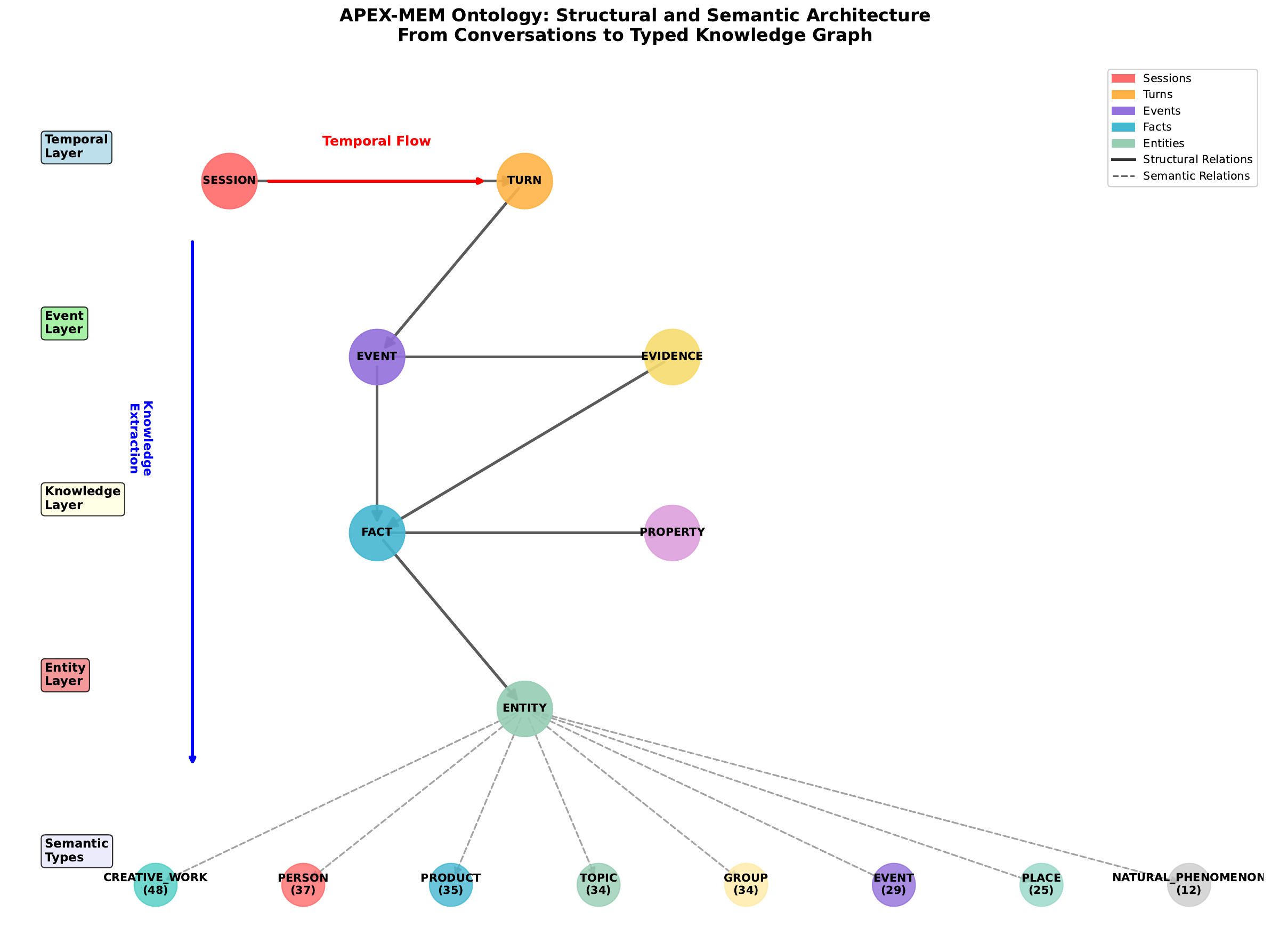}
\caption{APEX-MEM Ontological Architecture: Complete structural and semantic view showing the flow from conversational sessions through temporal events to typed knowledge extraction. The diagram illustrates both table-level relationships (solid arrows) and semantic type instantiation (dashed arrows), demonstrating how the system transforms unstructured conversations into a rich, ontology-grounded knowledge graph with diverse entity types including PERSON, CREATIVE\_WORK, PLACE, PRODUCT, and others.}
\label{fig:apex_mem_ontology}
\end{figure*}

Figure~\ref{fig:apex_mem_ontology} presents the complete ontological architecture of APEX-MEM, showing both the structural relationships between data components (sessions, turns, events, facts, entities) and the semantic taxonomy of entity types. This meta-level view illustrates how conversational data flows through the system's layered architecture to create a rich, typed knowledge graph.

The ontological architecture consists of five distinct layers that transform diverse information sources into structured knowledge:

\textbf{Temporal Layer:} The foundation layer captures the chronological structure of information flows from multiple sources. \textit{Sessions} represent complete interaction sequences with unique identifiers and timestamps, encompassing not only user-agent conversations but also news cycles, document streams, or event sequences. \textit{Turns} represent individual information instances within sessions, containing source, recipient, and content. For example, in conversational AI, turns capture user-agent exchanges; in news monitoring, turns represent individual news articles where "World" is the speaker and "LLM Agent" is the listener; in document processing, turns represent document ingestion events. This layer preserves temporal ordering essential for understanding information context and evolution across diverse domains.

\textbf{Event Layer:} This layer extracts meaningful temporal events from information turns across multiple domains. \textit{Events} represent semantically significant occurrences with structured metadata including event type, participants, location, and temporal anchors. Examples include conversational meetings, global news events (elections, natural disasters, policy changes), document publications, or system interactions. \textit{Evidence} provides the textual spans that support event extraction, maintaining traceability between structured representations and original content sources. This design enables APEX-MEM to process heterogeneous information streams—from intimate conversations to global news feeds—using the same architectural principles.

\textbf{Knowledge Layer:} The core semantic layer transforms events into structured knowledge regardless of information source. \textit{Facts} represent subject-property-value assertions extracted from events, with temporal validity intervals and confidence scores. These facts can describe personal preferences from conversations, geopolitical developments from news, or technical specifications from documents. \textit{Properties} define the semantic relationships and attributes that can be asserted about entities, with typed data schemas (string, boolean, date, etc.) that ensure consistency across diverse information domains and enable unified querying.

\textbf{Entity Layer:} This layer manages the canonical representation of all named entities mentioned across information sources. \textit{Entities} serve as the primary nodes in the knowledge graph, with unique identifiers, normalized names, and semantic type classifications. The entity resolution process ensures that mentions of the same entity across different contexts—whether in personal conversations, news articles, or documents—are properly linked and canonicalized, enabling cross-domain knowledge integration.

\textbf{Semantic Layer:} The top layer provides the ontological taxonomy that governs entity classification across all information domains. Entity types include \textit{PERSON} (individuals in conversations, public figures in news), \textit{CREATIVE\_WORK} (personal stories, published content), \textit{PLACE} (conversation locations, global regions), \textit{PRODUCT} (personal tools, commercial products), \textit{GROUP} (social circles, organizations, nations), \textit{TOPIC} (discussion subjects, news themes), \textit{EVENT} (personal activities, world events), and \textit{NATURAL\_PHENOMENON} (local weather, global climate events). This semantic typing enables sophisticated reasoning and querying that spans from personal context to global knowledge.

The knowledge extraction process (vertical flow) systematically transforms unstructured information from any source through these layers, while the temporal flow (horizontal) preserves chronological relationships essential for understanding information dynamics and evolution over time. This architecture's generality enables APEX-MEM to serve as a unified memory system for diverse AI applications, from personal assistants processing conversations to news analysis systems processing global events, all using the same ontological foundation.

\end{document}